\documentclass[10pt,twocolumn,letterpaper]{article}

\usepackage{wacv}              %

\usepackage{graphicx}
\usepackage{amsmath}
\usepackage{amssymb}
\usepackage{booktabs}

\usepackage{pifont}
\usepackage{tabularx}
\usepackage{relsize}
\usepackage{subcaption}
\usepackage{listings}
\usepackage{threeparttable}
\usepackage{ulem}
\usepackage[skip=0.1pt]{caption}
\usepackage{tikz}
\usetikzlibrary{tikzmark}
\usetikzlibrary{calc}
\usepackage{tikzpagenodes}
\usepackage{multirow}
\usepackage{booktabs}
\usepackage[super]{nth}
\usepackage{float}
\usepackage[accsupp]{axessibility}
\usepackage{balance}

\usepackage[pagebackref,breaklinks,colorlinks]{hyperref}

\usepackage[capitalize]{cleveref}
\crefname{section}{Sec.}{Secs.}
\Crefname{section}{Section}{Sections}
\Crefname{table}{Table}{Tables}
\crefname{table}{Tab.}{Tabs.}

\definecolor{somegray}{rgb}{0.5, 0.5, 0.5}
\newcommand{\darkgrayed}[1]{\textcolor{somegray}{#1}}
\makeatletter
\newcommand*\titleheader[1]{\gdef\@titleheader{#1}}
\AtBeginDocument{%
  \let\st@red@title\@title
  \def\@title{%
    \vskip-4em
    \bgroup\normalfont\large\centering\@titleheader\par\egroup
    \vskip1.5em\st@red@title}
}
\makeatother

\titleheader{\darkgrayed{This paper has been accepted for publication at the\\
IEEE Winter Conference on Applications of Computer Vision (WACV), Waikoloa, 2024.
\copyright IEEE}}

\def\wacvPaperID{2025} %
\def\confName{WACV}
\def\confYear{2024}

\title{Revisiting Token Pruning for Object Detection and Instance Segmentation}

\author{Yifei Liu
\quad
Mathias Gehrig
\quad
Nico Messikommer
\quad
Marco Cannici
\quad
Davide Scaramuzza\\
Robotics and Perception Group, University of Zurich, Switzerland\\
{\tt\small \{yifei.liu@, mgehrig@ifi., nmessi@ifi, cannici@ifi., sdavide@ifi.\}uzh.ch}
}

\begin{document}
\maketitle

\begin{abstract}
Vision Transformers (ViTs) have shown impressive performance in computer vision, but their high computational cost, quadratic in the number of tokens, limits their adoption in computation-constrained applications. However, this large number of tokens may not be necessary, as not all tokens are equally important. In this paper, we investigate token pruning to accelerate inference for object detection and instance segmentation, extending prior works from image classification. Through extensive experiments, we offer four insights for dense tasks: (\romannumeral 1) tokens should not be completely pruned and discarded, but rather preserved in the feature maps for later use. (\romannumeral 2) reactivating previously pruned tokens can further enhance model performance. (\romannumeral 3) a dynamic pruning rate based on images is better than a fixed pruning rate. (\romannumeral 4) a lightweight, 2-layer MLP can effectively prune tokens, achieving accuracy comparable with complex gating networks with a simpler design.  
We assess the effects of these design decisions on the COCO dataset and introduce an approach that incorporates these findings, showing a reduction in performance decline from $\sim$1.5 mAP to $\sim$0.3 mAP in both boxes and masks, compared to existing token pruning methods.
In relation to the dense counterpart that utilizes all tokens, our method realizes an increase in inference speed, achieving up to 34\% faster performance for the entire network and 46\% for the backbone.
\end{abstract}
\hspace{-7pt}
\vspace{-1pt}
\textbf{Code: }\url{https://github.com/uzh-rpg/svit/}
\section{Introduction}\label{sec:intro}

\begin{figure}[h]
\begin{center}
   \includegraphics[width=1.\linewidth]{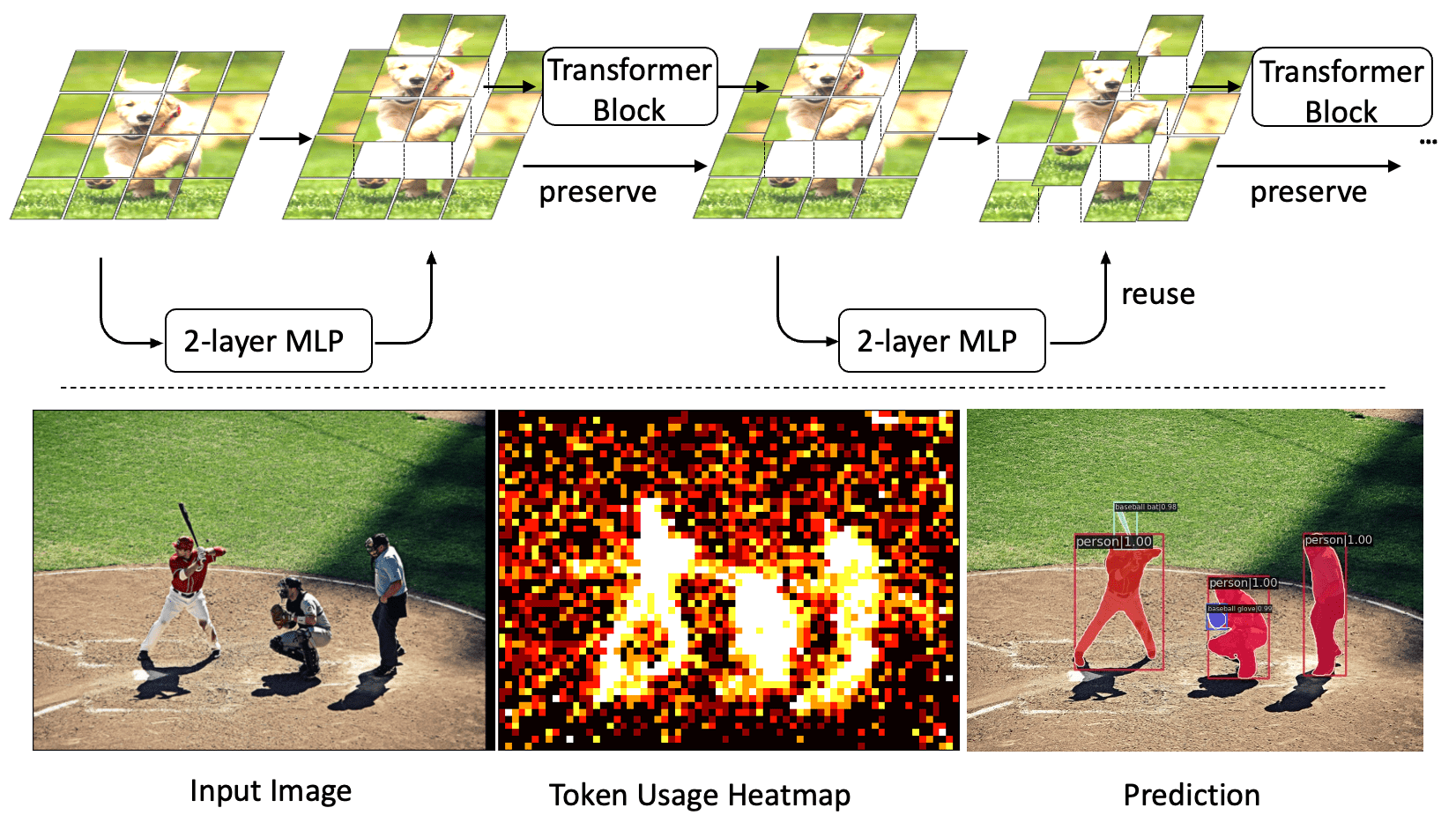}
\end{center}

   \caption{Top: high-level workflow of SViT, the MLP selectively chooses tokens to be processed in the transformer block, and the pruned tokens are preserved in feature maps and can be reactivated in later layers. Bottom: the token useage heatmap represents the number of layers using the tokens, and shows that the computational distribution highly aligns with interested objects.}
\label{fig:long}
\label{fig:eye_catching}
\end{figure}

    \begin{figure*}[h]
        \begin{center}
        \includegraphics[width=0.9\linewidth]{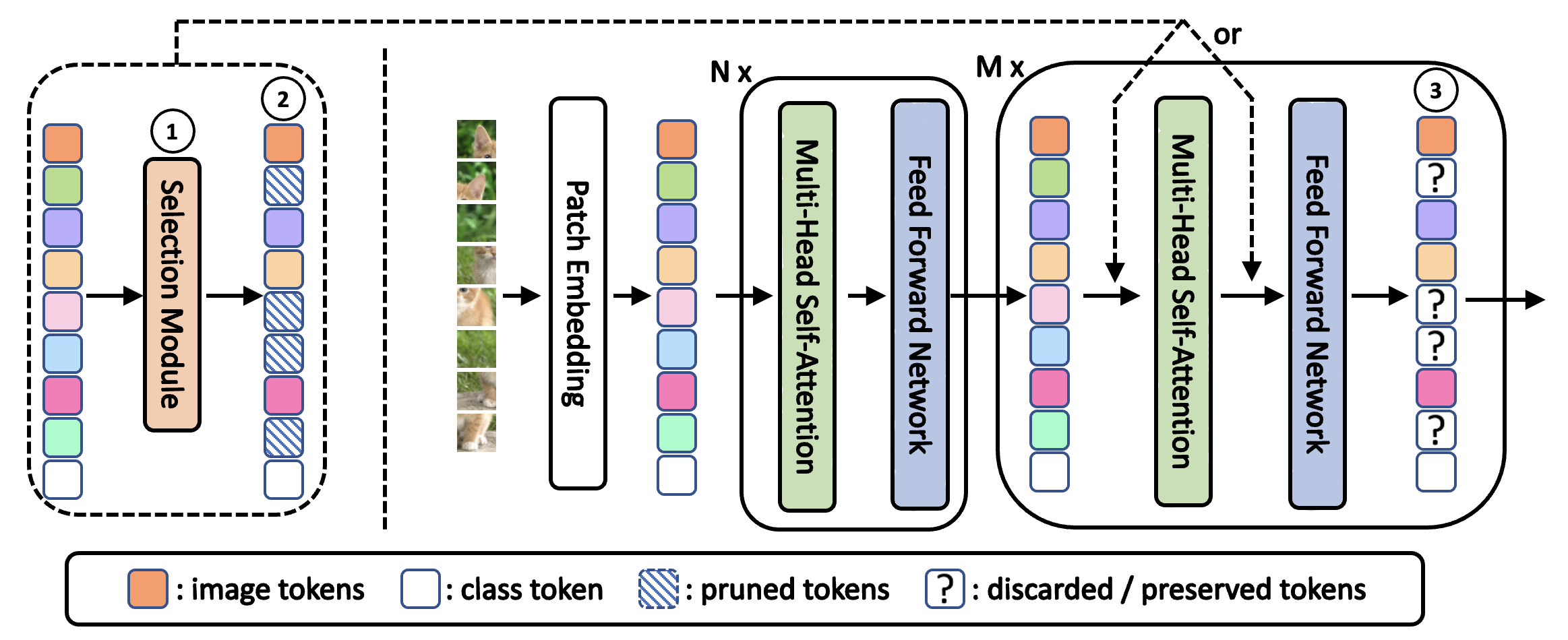}
        \end{center}

       \caption{A high-level comparison of the overall workflows for various token pruning methods. \ding{172} Selection Module: may utilize a gating module (before self-attention) or be attention-based (after self-attention). \ding{173} Number of pruned tokens: can be either dynamic or fixed. \ding{174} Treatment of pruned tokens: either removing or preserving them. If they are preserved within feature maps, there is an additional option to reactivate them.}
        \label{fig:overview}
    \end{figure*}

Transformers and multi-head self-attention \cite{AttentionIsAllYouNeed} have revolutionized the field of computer vision.
Since their first introduction, Vision Transformers (ViTs) \cite{ViT,DeiT,Swin} have quickly become the leading model architecture for a number of vision tasks, including image classification \cite{DeiT, levit, Swin}, object detection \cite{Swin, detr, vitadapter, exploring_plain_for_detection, zhu2020deformable}, semantic segmentation \cite{cheng2021per,zheng2021rethinking,Swin,xie2021segformer}, and others \cite{han2022survey,khan2022transformers}.
Their unique ability to perform global reasoning through pair-wise token attention is, however, both a strength and a weakness.
Although it enhances the representational power of these architectures, it also leads to a significant increase in computational footprint, limiting the adoption of ViTs in resource-constrained settings.

A viable strategy for mitigating the substantial computational demands involves leveraging input-aware inference to prune less critical features within the input space. 
While this strategy has previously been applied to CNNs \cite{figurnov2017spatially}, resulting in improved FLOP measurements, the intrinsic regularity of convolution operations makes it difficult to obtain noticeable speedup on hardware.
However, the advent of ViTs paves the way for input-space pruning, 
as the MLPs in ViTs operate pointwise and self-attention inherently accommodates an arbitrary number of tokens. 
Consequently, pruning tokens can readily attain remarkable speedup without necessitating any additional hardware adaptations.

Initial investigations in the domain of token pruning have encompassed the utilization of gating networks to identify less significant tokens \cite{DynamicViT, SPViT, AdaViT} or eliminating tokens receiving minimal attention from the class token \cite{evit, ATS, EvoViT}.
These approaches, while having demonstrated their effectiveness,
were only applied to classification and
have yet to be applied to other tasks such as object detection and instance segmentation.
To the best of our knowledge, the exploration of token pruning in the context of dense tasks remains still notably scarce (Section \ref{sec:related_works})\footnote{The most related work on dense tasks is an extended version \cite{rao2023dynamic} of DynamicViT. However, it focuses on skipping MLPs in hierarchical models for dense tasks.}.
In this paper, we investigate token pruning for object detection and instance segmentation on isotropic vision transformers, with the aim of bridging the gap between classification and dense tasks.
During our preliminary experiments, we adapted prior methods to dense tasks and discovered they have apparent performance loss (Section \ref{subsec:evaluation of insights}).
With extensive experiments, we identified four key insights that are beneficial for improving model performance and simplifying model designs (Section \ref{subsec:insights_and_observations}), 
leading to a method that outperforms previous state-of-the-art token pruning methods by a significant margin on object detection and instance segmentation (Section \ref{subsec: comparison_state_of_the_art}). Our insights are as follows:

\textbf{Token preserving on dense tasks.}
Unlike classification, where pruned tokens can be removed permanently, dense prediction tasks benefit from preserving them in feature maps for subsequent utilization by the detection head.

\textbf{Token reactivation as needed.}
In addition to preserving them, reactivating pruned tokens in the backbone on demand can improve model performance by adapting to layer-wise attention and recovering mis-pruned tokens for better robustness. A token once pruned has the flexibility to be reused at any subsequent layer, including the immediately succeeding one.

\textbf{Pruning with a dynamic rate.}
The concept of a dynamic pruning rate, previously introduced for classification tasks in \cite{A-ViT, SPViT, ATS}, optimizes model performance within the same computation resource by allocating more tokens for complex images and fewer for simple images. It gains additional efficacy when integrated with token reactivation on dense prediciton tasks.

\textbf{2-layer MLP is sufficient.}
A lightweight MLP is sufficient to select which tokens should be pruned, delivering almost the same accuracy as more complex gating networks \cite{DynamicViT, SPViT} used for classification.

We evaluate these design choices and build upon them to introduce a straightforward model to selectively prune tokens,  which we refer to as SViT.
We demonstrate that this model surpasses previous state-of-the-art token pruning models by reducing loss in mAP from $\sim$1.5 to $\sim$0.3 for both boxes and masks, and accelerates the inference speed of the dense counterpart by up to 34\% for the whole network and 46\% for the backbone.

\section{Related Work}
\label{sec:related_works}
\textbf{Vision\hspace{0.05cm} Transformer}\hspace{0.25cm}
Originating in the NLP community \cite{devlin2018bert,liu2019roberta,radford2018improving,brown2020language}, Transformers \cite{AttentionIsAllYouNeed} have lately acquired popularity also in the field of computer vision for their ability to capture long-range relations \cite{khan2022transformers,han2022survey}. 
The seminal work on Vision Transformers (ViTs) \cite{ViT} demonstrated state-of-the-art classification performance, when pre-trained on large-scale datasets.

Since then, several improvements have been proposed to the ViT architecture, including improved tokens' aggregation schemes \cite{T2T,DeiT,TinT,AllTokensMatter}, multi-scale hierarchical designs \cite{PyramidViT,Pvtv2,Twins,Swin,Regionvit,CrossFormer,CrossFormerPP}, and hybrid architectures combining CNNs \cite{CeiT,ResT,NesT} 
. 
Apart from design improvements, researchers have also investigated their use in more complex vision tasks \cite{Swin,vitadapter,exploring_plain_for_detection,zheng20213d,yu2021pointr,zhao2021point}.
This paper fits in between these two lines of research, as we not only focus on  architecture design choices, but also extend their usage to dense prediction tasks such as object detection and instance segmentation.

\textbf{Transformer\hspace{0.05cm} Acceleration}\hspace{0.25cm}
Various methods have been explored for optimizing Transformers' high computational cost, including designing alternative lightweight attention formulations \cite{kitaev2020reformer,child2019generating,katharopoulos2020transformers,roy2021efficient, tay2020sparse, vyas2020fast, zaheer2020big}, removing unnecessary network modules \cite{michel2019sixteen,voita-etal-2019-analyzing,fan2019reducing} approximating attention multiplications with low-rank decompositions \cite{chen2021scatterbrain,choromanski2020rethinking,wang2020linformer},  distilling knowledge into a more efficient student network \cite{DeiT,sanh2019distilbert,zhang2022minivit}, and extending network quantization techniques for Transformers \cite{zafrir2019q8bert,fan2020training,shen2020q,kim2021bert, bhandare2019efficient}. Furthermore, acceleration techniques specific to ViTs have been proposed \cite{tokenlearner,DynamicViT,evit,EvoViT,A-ViT,ATS, IA-RED2} by exploiting the redundancy in the input patches to early drop tokens for saving computation.

\textbf{Input\hspace{0.25cm}Space\hspace{0.25cm}Pruning}\hspace{0.25cm}
As not all regions in the input image are equally important, pruning redundant areas can save computation without apparent accuracy loss. Spatially ACT \cite{figurnov2017spatially} prunes pixels for CNNs. 
Numerous token pruning methods for ViTs have been developed on classification, including using gating networks \cite{DynamicViT, SPViT, AdaViT}, attention scores \cite{evit, EvoViT, ATS}, reinforcement learning \cite{IA-RED2} and others \cite{A-ViT, tokenlearner, ToMe}. Among them, ToMe \cite{ToMe} proposes to merge tokens rather than remove them. A few works also consider dense tasks: SparseViT \cite{sparsevit} prunes coarse windows for pyramid transformers, while we prune finer-grained tokens for isotropic transformers. SparseDETR \cite{sparseDETR} focuses on improving the efficiency of DETR \cite{detr} architecture, while we focus on improving transformer-based backbones. STViT-R \cite{STViT} sparsifies tokens by repeatedly clustering them into a few semantic tokens and restoring the spatial resolution, while we keep the spatial resolution with detailed position information.

\section{Token Pruning on dense prediction tasks}
\label{sec:selective_vision_transformer}

\subsection{Revisit prior token pruning approaches}
We review the majority of token pruning techniques by illustrating the high-level distinctions in their workflows.
 As shown in Table \ref{tab:relation_of_methods}, these approaches can be classified along four dimensions: the selection module, use of dynamic pruning rate, preservation of pruned tokens, and reactivation of pruned tokens.

The overall workflow of token pruning is depicted in Figure \ref{fig:overview} and can be summarized as follows: initially, the input image is partitioned into non-overlapping patches, which are linearly transformed into tokens and subsequently processed by the initial ViT blocks to obtain comprehensive enough feature representations. Then, token selection modules are introduced to identify tokens for pruning, consequently accelerating computations due to the reduced number of tokens. Note that, here, acceleration comes out-of-the-box as self-attention can adaptively process fewer number of tokens without any modification.

\begin{table*}
\begin{center}
\begin{threeparttable}
\begin{tabular}{c c c c l}
\hline
Selection Module
&
\begin{tabular}{@{}c@{}}Dynamic \\ Pruning \\ Rate \end{tabular}
&
\begin{tabular}{@{}c@{}}
Preserve \\ Pruned \\ Tokens\end{tabular}
&
\begin{tabular}{@{}c@{}}
Reactivate \\ Pruned \\ Tokens\end{tabular}
&
Model\\
\hline\hline
gating module & \ding{51} & \ding{51} & \ding{51} &SViT (Ours)\\
gating module & \ding{55} & \ding{55} & \ding{55} &DynamicViT \cite{DynamicViT}\\
attention-based & \ding{51} & \ding{55} & \ding{55} &ATS \cite{ATS}\\
attention-based & \ding{55} & \ding{51} & \ding{55}\tnote{2} &Evo-ViT \cite{EvoViT}\\
attention-based & \ding{55} & \ding{55} & \ding{55} &EViT \cite{evit}\\

\hline
\end{tabular}
\begin{tablenotes}
\item [\scriptsize 2] \footnotesize While Evo-ViT is theoretically capable of reusing tokens by design, it tends to use the same tokens throughout the network, details in the supplementary material.
\end{tablenotes}
\end{threeparttable}
\end{center}
\caption{A high-level examination of token pruning techniques. The gating module refers to an auxiliary compact network, designed to predict the tokens to be pruned. Attention-based selection involves pruning tokens that receive minimal attention from the class token.}
\label{tab:relation_of_methods}
\end{table*}

\subsection{Insights and Observations} \label{subsec:insights_and_observations}
\paragraph{Preserve pruned tokens within feature maps.}
A notable distinction between classification and dense prediction tasks is how the pruned tokens should be treated. 
In classification, token pruning methods often remove tokens permanently because pruned tokens will no longer influence the result, as the classification solely depends on the class token, which is always kept. 

However, on dense prediction tasks, the pruned tokens can still be utilized by subsequent detection heads, even if they are no longer updated in the backbone. Therefore, it is beneficial to keep the already computed features for pruned tokens for later use.
When pruned tokens are not preserved, we recover a dense feature map by placing remaining tokens in their original location, and zero-pad the pruned ones\cite{ye2022ostrack}. Preserving pruned tokes, instead, built the feature map incrementally, each time replacing updated tokens, but keeping pruned ones unchanged.
Preserving pruned tokens can be as fast as removing them (see Table \ref{table:effect_preserve}), and improves model performance on various models on dense tasks.

\paragraph{Reuse preserved tokens on demand.}
As pruned tokens are preserved within feature maps, it is natural to consider whether they should be used again. 
In the scope of this paper, "token preserving" refers to the utilization of pruned tokens only by detection heads, whereas "token reactivation" implies that these tokens can also be reintroduced into the backbone for subsequent layers.
A counterargument to token reactivation may say that ViT should prioritize allocating computing resources to informative tokens as much as possible \cite{EvoViT}, and reactivating pruned tokens may potentially undermine this principle.
However, the definition of "informative" may vary across different layers since ViT could concentrate on distinct regions at each layer, see supplementary material.
Thus, the ability to reactivate pruned tokens accommodates the distinct attentions of various ViT layers, enabling the model to prioritize its current focus before returning to other relevant tokens in subsequent blocks. Additionally, this makes pruning more robust, as mis-pruned tokens have the opportunity to become active again, see Figure \ref{subfig:reactivation_example}. Ultimately, these advantages lead to a more effective overall utilization of tokens under the same token usage per block. In Section \ref{sec:experiments}, we allow the model to learn whether and when to reuse pruned tokens by itself, and show that this ability can improve model accuracy by a 0.4 box AP and 0.3 mask AP.
\begin{figure*}[!htb]
    \centering
    \begin{subfigure}[b]{0.247\textwidth}
        \includegraphics[width=\textwidth]{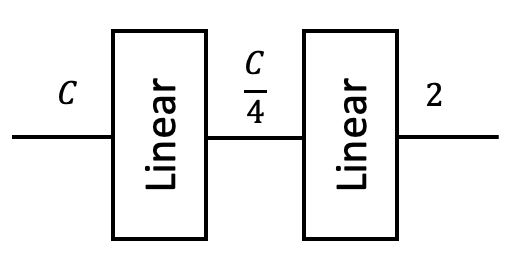}

        \caption{Two-layer gating network}

        \label{fig:subfig_2layerMLP}
    \end{subfigure}
    \hspace{10pt}
    \begin{subfigure}[b]{0.52\textwidth}
        \includegraphics[width=\textwidth]{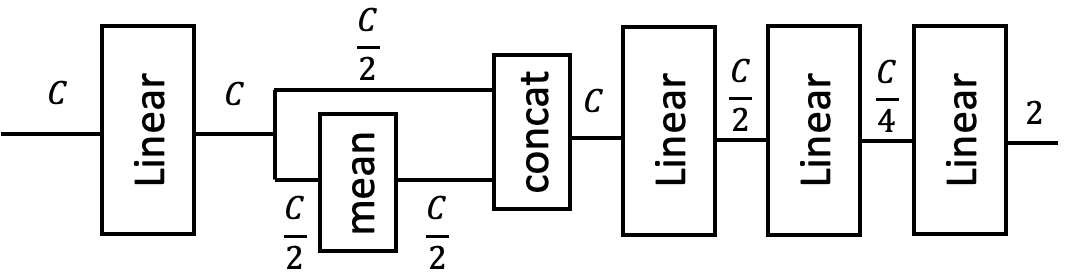}
        \caption{Gating network used in DynamicViT}
        \label{fig:subfig_complex}
    \end{subfigure}
    \caption{Different types of gating networks for predicting tokens to be pruned. (a) is used by SViT, and (b) is used by DynamicViT~\cite{DynamicViT}. Normalization and activation functions are omitted for conciseness. $C$ represents the dimension of tokens.}
    \label{fig:gating_design}
\end{figure*}

\paragraph{A 2-layer MLP can substitute complex gating networks for pruning tokens.}
Prior token pruning approaches tend to employ complex gating networks for predicting the tokens to be pruned. 
In DynamicViT \cite{DynamicViT}, several MLPs are utilized in conjunction with mean and concatenation operations to learn both token-specific and global information for determining which tokens should be pruned, as illustrated in Figure \ref{fig:subfig_complex}. 
SPViT \cite{SPViT}, introduces a more intricate gating network that incorporates an additional head branch to calculate score weights for each individual head. 
However, in Section \ref{subsec:evaluation of insights}, our study shows that a simple 2-layer MLP in Figure \ref{fig:subfig_2layerMLP} works equally well and simplifies the architecture design.

\paragraph{A dynamic pruning rate is better than a fixed pruning rate.}
Several studies \cite{A-ViT, SPViT, ATS} in the context of classification have implemented dynamic pruning rates, adaptively pruning varying numbers of tokens based on the input images during inference.
We further validate its effectiveness in the context of object detection and instance segmentation, and show it is one key components to achieve optimal performance in Section \ref{sec:experiments}.\\

\subsection{SViT: Selective Vision Transformer}

In light of the insights, we introduce the Selective Vision Transformer (SViT), a simple yet effective token pruning model, which seamlessly integrates all prior findings.

SViT is depicted in Figure \ref{fig:eye_catching}. For the selection module, we employ a 2-layer perceptron followed by Gumbel Softmax \cite{gumbel1, gumbel2} to make the discrete decision differentiable, as shown in Eq (\ref{eq: gating_network}). By placing this selection module before the entire ViT block, we facilitate acceleration for both self-attention and the MLP in the transformer encoder:

    \begin{equation}
        \begin{aligned}
        &\mathbf{p} = \text{Softmax}(\text{MLP}(\mathbf{x})) \in \mathbb{R}^{N \times 2}, \
        \mathbf{x} \in \mathbb{R}^{N \times C}\\
        &\mathbf{M} = \text{GumbelSoftmax}(\mathbf{p}) \in \{0, 1\}^{N},\\
        &\mathbf{x} \leftarrow \mathbf{M} \odot \text{ViTBlock}(\mathbf{x}, \mathbf{M}) + (1 - \mathbf{M}) \odot \mathbf{x}
        \end{aligned}
        \label{eq: gating_network}
    \end{equation}
where $\mathbf{x}$ represents the input tokens, $\mathbf{p}$ is the intermediate sampling probability, $\mathbf{M}$ signifies token masks, and $\odot$ is Hadamard product. The MLP transforms token dimensions from $C$ to $\frac{C}{4}$, and $\frac{C}{4}$ to $2$.
The ViTBlock takes in the masks $\mathbf{M}$ and eliminates the influence of pruned tokens on other tokens during training by setting the corresponding columns in the attention matrix to 0.
During inference, we simply gather the active tokens, feed them to the current ViT Block, and then scatter them back to the previous feature map.

\begin{figure*}[h]
\begin{center}
\includegraphics[width=1\linewidth]{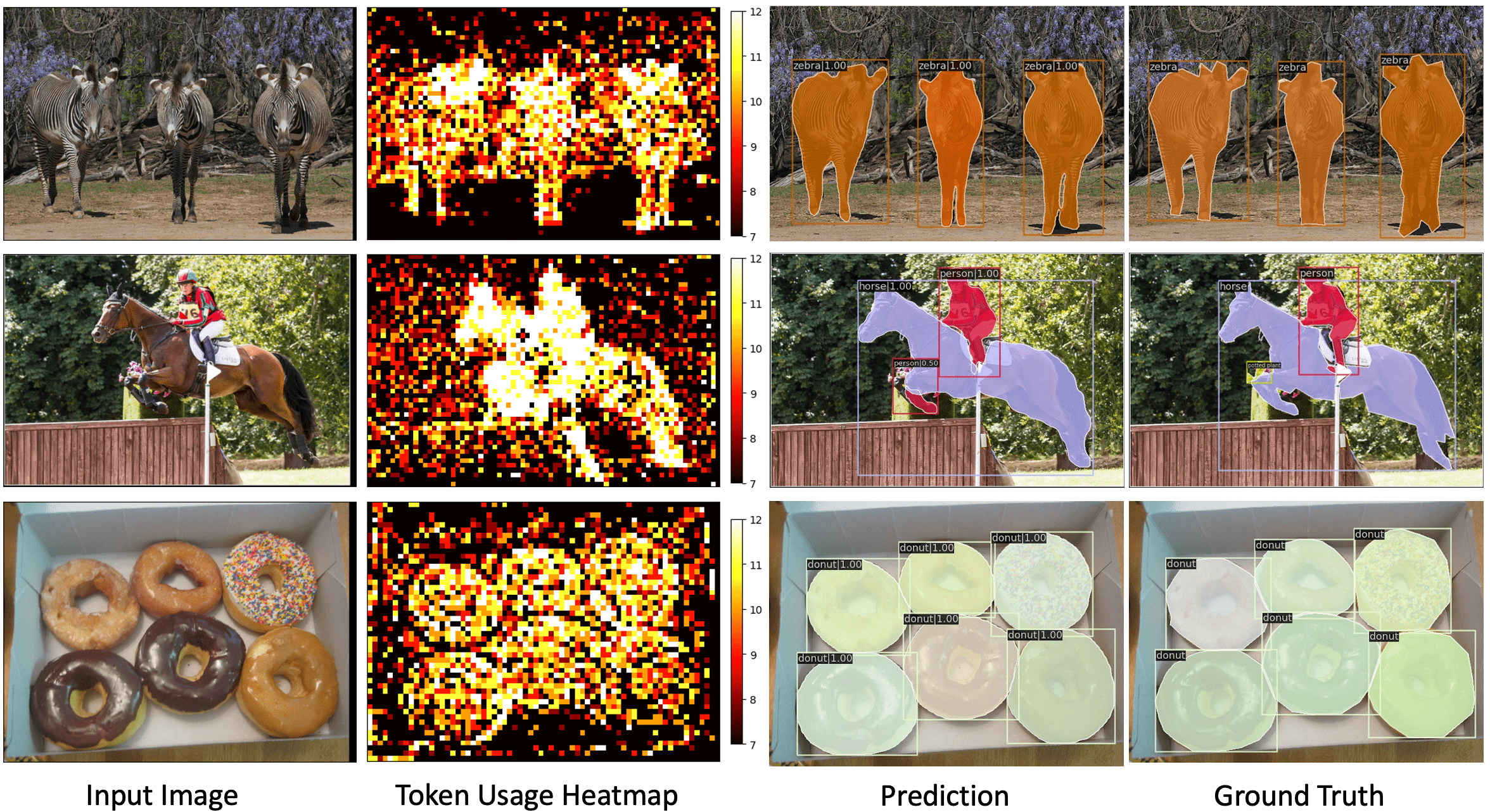}
\end{center}

   \caption{SViT learns to allocate computation to visually more important tokens. The token usage heatmap shows the number of  layers used for each token and reflects computational distribution over the input space, which highly align with fine-grained object contours. More visualizations and dataset-level statistics are in the supplementary material.}
\label{fig:qualitative_results}
\end{figure*}

For controlling the number of pruned tokens, similar to \cite{SPViT}, we use a dynamic pruning ratio loss during training as in Eq (\ref{eq:dynamic_ratio_loss}):

    \begin{equation}
        \begin{aligned}
        &\mathcal{L}_{dynamic} = \frac{1}{L}\sum_{l\in L}((\frac{1}{BN}\sum_{b \in B}\sum_{n \in N} {\mathbf{M}}_n^{b,l}) - \mathbf{t}^l)^2,\\
        &\rm{\mathcal{L}_{total} = \mathcal{L}_{task} + \lambda \mathcal{L}_{dynamic}}
        \end{aligned}
        \label{eq:dynamic_ratio_loss}
    \end{equation}
where $\mathbf{M}_n^{b,l}$ denotes the mask at batch $b$ and layer $l$ for the $n$-th token, $\mathbf{t}^l$ represents the target keeping ratio at layer $l$, and $\lambda$ is a hyper-parameter to weight losses. It is worth noting that the token usage, i.e averaged mask values: $\frac{1}{BN}\sum_{b \in B}\sum_{n \in N} {\mathbf{M}}_n^{b,l}$, is averaged not only across all tokens but also across images in a batch, making the loss aware of the trade-off between token usage and accuracy, resulting in more tokens allocated for complex images and fewer tokens for simpler images. For a comparison with a fixed pruning ratio loss, see Section \ref{subsec:evaluation of insights}.

\section{Experiments}
\label{sec:experiments}

We conduct experiments on the COCO 2017 object detection and instance segmentation dataset \cite{COCO}, which consists of 118K training images and 5K validation images, and provide experiments on ImageNet-1K \cite{ImageNet1k} classification in the supplementary material A.
We use Mask R-CNN \cite{Mask-RCNN} as our object detection framework, and  employ ViT-Adapter \cite{vitadapter} to wrap a ViT as the backbone.
The dense backbone utilizes DeiT \cite{DeiT} with global self-attention, while the sparse backbone adopts one of the token pruning models (DynamicViT, EViT, EvoViT, ATS, SViT) with a reduced number of tokens.
By default, SViT incorporates nine gating modules, ranging from the 4-th to the 12-th layer to prune tokens from the dense model, and adheres to the target keeping ratio of [70\%, 70\%, 70\%, 49\%, 49\%, 49\%, 34.3\%, 34.3\%, 34.3\%] following conventions \cite{DynamicViT, evit}.
For training, we follow the settings of ViT-Adapter \cite{vitadapter} to train the dense model with a 3x schedule (36 epochs). Then we finetune each sparse model for 6 and 4 epochs for tiny and small models, respectively, with an initial learning rate of 1e-5 and the loss hyper-parameter $\lambda=4$ .
In the following, we first present experiments for each insight, and then compare the derived SViT with other state-of-the-art token pruning models on object detection and instance segmentation. Finally, we analyse the pattern of pruning and reactivation by providing qualitative and quantitative results.

\subsection{Evaluation of the insights and observations}\label{subsec:evaluation of insights}
\paragraph{Preserve pruned tokens within feature maps}
We evaluate the difference between removing and preserving pruned tokens on four state-of-the-art models: EViT \cite{evit}, Evo-ViT \cite{EvoViT}, DynamicViT \cite{DynamicViT} and ATS \cite{ATS}. Some of these models prune tokens via the attention score from the class token, which does not naturally exist on dense tasks, and we insert an artificial class token and find it still works well for these models. As shown in Table \ref{table:effect_preserve}, Evo-ViT, which inherently preserves pruned tokens, performs the best among the original models. In addition, by enabling preserving tokens, EViT and ATS both get small increase in performance, and DynamicViT has a boost increase, as gating networks learned end-to-end are sensitive to gradient information kept in pruned tokens, and the gradients cannot be back-propagated to the backbone if pruned tokens are dropped. Owing to this factor, DynamicViT-S experiences training divergence, as indicated in \ref{tab:Performance_COCO}.
    \begin{table*}[ht]
    \begin{center}
        \begin{tabular}{l c c c c c c c} 
        \hline
          & \multicolumn{2}{c}{DynamicViT} & \multicolumn{2}{c}{EViT} & \multicolumn{2}{c}{ATS} & \multirow{2}{*}{Evo-ViT} \\
          \cmidrule(r){2-3} \cmidrule(lr){4-5} \cmidrule(lr){6-7}
          &  remove & prsv.& remove & prsv. & remove & prsv. & \\
        \hline
         AP\textsuperscript{box} & 41.2  & 44.1 & 44.5 & 44.7  & 43.9 & 44.1 & 44.8\\
         AP\textsuperscript{mask} & 37.1  & 39.3 & 39.8 & 39.9 & 39.1 & 39.3 & 39.9\\
         FPS &  23.10 & 22.95 & 22.76 & 22.81 & 16.41 & 16.52 & 22.12\\
        \hline
        \end{tabular}
    \end{center}

    \caption{Effectiveness of removing tokens vs. preserving tokens on COCO 2017. Evo-ViT inherently preserves tokens, and performs the best among these models; Similarly, preserving tokens in feature maps increases the performance of the other three models. As anticipated, the token removal or preservation process has a negligible impact on inference speeds; scattering updated tokens onto either a zero feature map or the previous feature map consumes equivalent computational time.}
    \label{table:effect_preserve}
    \end{table*}

\begin{figure*}[!htb]
    \centering
    \begin{subfigure}[b]{0.4\textwidth}
        \includegraphics[width=\textwidth]{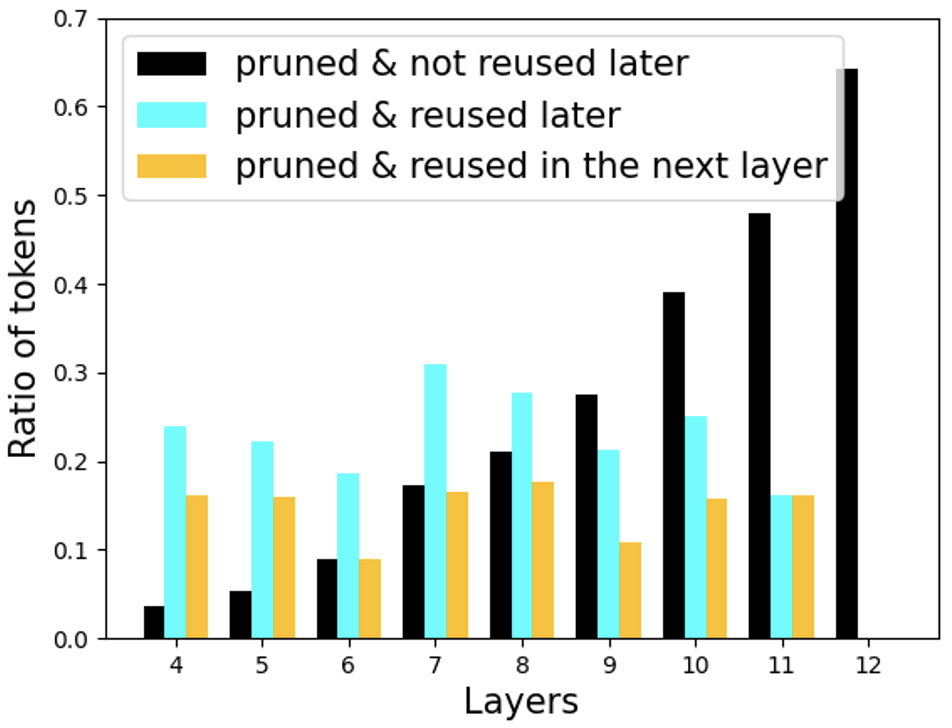}
        \caption{reactivation ratio per layer}
        \label{subfig:reactivation_ratio}
    \end{subfigure}
    \hspace{10pt}
    \begin{subfigure}[b]{0.55\textwidth}
        \includegraphics[width=\textwidth]{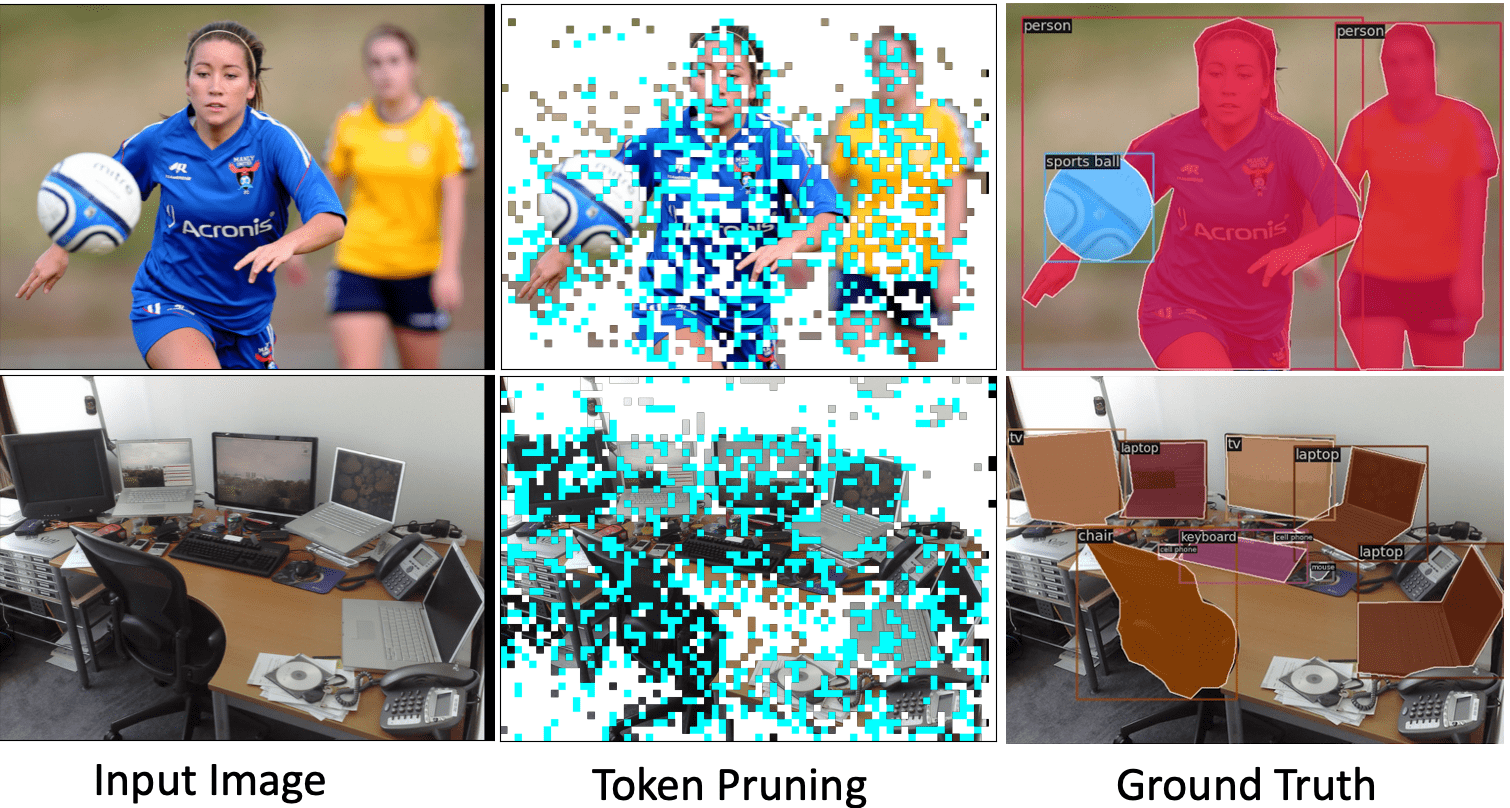}
        \caption{reactivation example}
        \label{subfig:reactivation_example}
    \end{subfigure}

    \caption{(a) Reactivation ratio at different layers, averaged on COCO validation set. (b) Visualization of reactivated tokens. Cyan tokens will be reactivated in later layers, while white tokens are not. Reactivated tokens are visually more important tokens.}
    \label{fig: reactivation}
\end{figure*}

\paragraph{Reuse preserved tokens at demand}
We evaluate the influence of reusing / reactivating  pruned tokens on SViT instead of the previous models, as models utilizing attention-based selection cannot really reuse tokens. Attention-based selection happens after Multi-Head Self-Attention (MHSA), and reusing tokens in such case requires all tokens to participate in MHSA, leading to no computational savings.  
To construct our baseline that is restricted not to reuse pruned tokens, we multiply the mask at $l$-th layer by its previous mask at $l-1$-th layer following \cite{DynamicViT}: $\mathbf{M}^l \leftarrow \mathbf{M}^l \odot \mathbf{M}^{l-1}$.
This implies that active tokens will consistently be a subset of previous active tokens, and pruned tokens cannot be used again. As the set of active tokens is strictly decreasing, we merge selection modules with the same keeping ratios into one selection module.
Table \ref{tab:Ablation_dynamic_&_preserve} shows that by reusing pruned tokens, SViT-T gets +0.4 box AP and +0.3 mask AP. Reactivation ratio and visualizations samples are in Figure \ref{fig: reactivation}.
    \begin{table}
        \begin{center}
        \begin{tabular}{@{}l c c c c@{}}
        \hline
        Model  &  dynamic & reactivation &
        AP\textsuperscript{box} & AP\textsuperscript{mask} \\
        \hline
         SViT-T  & \ding{51} & \ding{51}   &  45.5 &  40.7 \\ 
         SViT-T  & \ding{55} & \ding{51}   &  45.1 &  40.2 \\
         SViT-T  & \ding{51} & \ding{55}   &  45.1 &  40.4 \\
         SViT-T  & \ding{55} & \ding{55}   &  44.9 &  40.2 \\ 
        \hline
        \end{tabular}
        \end{center}
        \vspace{-5pt}
        \caption{The effects of dynamic pruning rate and reactivating pruned tokens. Both can enhance performance individually, and their combination results in a larger improvement. }
        \label{tab:Ablation_dynamic_&_preserve}
    \end{table}

\paragraph{Dynamic pruning rate outperforms fixed pruning rate}
To evaluate the influence of dynamic pruning rate vs. fixed pruning rate, we create a baseline by changing the dynamic ratio loss $\mathcal{L}_{dynamic}$ from equation \eqref{eq:dynamic_ratio_loss} to the fixed ratio loss $\mathcal{L}_{fixed}$ \cite{DynamicViT} as follows:
    \begin{equation}
        \mathcal{L}_{fixed} = \frac{1}{LB}\sum_{l\in L}\sum_{b \in B}\left(\left(\frac{1}{N}\sum_{n \in N} {\mathbf{M}}_{b,n}^l\right) - \mathbf{t}^l\right)^2,
        \label{eq:fixed_ratio_loss}
    \end{equation}

This loss does not average token usage across images within a batch, thereby penalizing each image towards the same keeping ratio. As indicated in Table \ref{tab:Ablation_dynamic_&_preserve}, employing just a dynamic pruning rate yields a gain of +0.2 in both box AP and mask AP for SViT-T. When further augmented with token reactivation, these improvements escalate to +0.4 for box AP and +0.5 for mask AP. This substantiates both the efficacy of implementing a dynamic pruning rate in dense tasks and the added benefits of its integration with token reactivation. We also provide throughput experiments with different batch sizes in the supplementary material.

\paragraph{A 2-layer MLP performs as good as complex gating networks}
We evaluate the designs of the different gating modules, as shown in Figure \ref{fig:gating_design}, and experiment with both tiny and small models. Table \ref{tab:Ablation_gating_design} shows that using a 2-layer MLP to predict tokens for pruning achieves the same box AP and only -0.1 mask AP for tiny models, and -0.1 box AP and the same mask AP for small models. This verifies the role of 2-layer MLP as an effective and simple selection module.

    \begin{table}
        \begin{center}
        \begin{tabular}{@{}l c c c c@{}}
        \hline
        \multirow{2}{*}{\vspace{-5pt}Gating module}  & \multicolumn{2}{c}{Tiny} & \multicolumn{2}{c}{Small} \\
        \cmidrule(lr){2-3} \cmidrule(l){4-5}
         & AP\textsuperscript{box} & AP\textsuperscript{mask} & AP\textsuperscript{box} & AP\textsuperscript{mask}\\
        \hline
         2-layer MLP    &  48.2 &  48.5 & 45.5 & 40.7\\ 
         complex gating network    &  48.2 &  48.6 & 45.6 & 40.7\\
        \hline
        \end{tabular}
        
        \end{center}
        \vspace{-5pt}
        \caption{Evaluation of designs for the gating module on SViT-T and SViT-S. A simple MLP can achieve similar performance with complex gating network, simplifying model design.}
        \vspace{-10pt}
        \label{tab:Ablation_gating_design}
    \end{table}

\subsection{Comparison with state-of-the-art models} \label{subsec: comparison_state_of_the_art}
    \begin{table*}
    \setlength{\tabcolsep}{4pt}
        \begin{center}
        \begin{tabular}{l c c c c c c c c}
        \hline
        \multirow{2}{*}{\shortstack{ \vspace{2pt} \\ Model in \vspace{3pt} \\ ViT-Adapter}} & \multicolumn{4}{c}{Tiny} & \multicolumn{4}{c}{Small} \\
        \cmidrule(r){2-5} \cmidrule(l){6-9}
          & AP\textsuperscript{box} & AP\textsuperscript{mask} & FPS\textsuperscript{w} & FPS\textsuperscript{b} & AP\textsuperscript{box} & AP\textsuperscript{mask} & FPS\textsuperscript{w} & FPS\textsuperscript{b}\\
        \cmidrule(r){1-1} \cmidrule(r){2-5} \cmidrule(l){6-9}
        DeiT \cite{DeiT} & 45.8 & 40.9 & 18.45 & 27.61 & 48.5 & 42.8 & 11.70 & 14.20\\ 
        \cmidrule(r){1-1} \cmidrule(r){2-5} \cmidrule(l){6-9}
        EViT \cite{evit}  & 44.5 (-1.3) & 39.8 (-1.1) & 22.76 & 35.80 & 47.1 (-1.4) & 41.6 (-1.2) & 15.34 & 20.01\\ 
        EvoViT \cite{EvoViT}& 44.8 (-1.0) & 39.9 (-1.0) & 22.12 & 34.33& 47.2 (-1.3)& 41.6 (-1.2) & 15.48 & 20.26\\
        ATS \cite{ATS}   & 43.9 (-1.9) & 39.1 (-1.8) & 16.41 &22.38 & 46.7 (-1.8) & 41.1 (-1.7) & 11.63 & 14.24\\
        DyViT \cite{DynamicViT} & 41.2 (-4.6) & 37.1 (-3.8) & \textbf{23.10} &\textbf{36.45} & diverge & diverge & / &/\\
        DyViT+prsv. & 44.1 (-1.7) & 39.3 (-1.6) & 22.95 & 36.38& 47.2 (-1.3) & 41.6 (-1.2) & 15.66 & \textbf{20.79}\\
        SViT (Ours) & \textbf{45.5 (-0.3)} & \textbf{40.7 (-0.2)} & 22.32 &34.69 & \textbf{48.2 (-0.3)} & \textbf{42.5 (-0.3)} &  \textbf{15.75} & 20.78\\
        \hline
        \end{tabular}
        \end{center}
        \vspace{-5pt}
        \caption{Comparison of token pruning methods on COCO object detection and instance segmentation. DeiT is the dense model using all tokens. FPS\textsuperscript{w} and FPS\textsuperscript{b} represents the inference speeds for the whole network and the backbone, respectively, which are measured with batch size 1 on a single A100 GPU.}
        \label{tab:Performance_COCO}
        \vspace{-5pt}
    \end{table*}

In this section, we compare SViT with prior art pruning models adapted for dense tasks. We evaluate inference speeds on a NVIDIA A100 GPU for both the backbones and the entire networks. As illustrated in Table \ref{tab:Performance_COCO}, sparse models exhibit comparable relative speed gains under identical total pruning rates, with the exception of ATS. The latter incurs computational overhead in its inverse transform sampling module for dealing with a large number of tokens in dense tasks. Among sparse models, SViT gets the highest performance for both tiny and small models. SViT-S significantly surpasses all baseline models, narrowing the performance drop with respect to the dense model from a range of -1.3 to -1.8 in box AP and -1.2 to -1.7 in mask AP, down to a mere -0.3 for both metrics. This performance advantage is consistently observed in SViT-T as well. In comparison with the dense counterpart, SViT-S improves inference speed by $\sim$34\% and $\sim$46\% for the entire network and the backbone, respectively, with negligible -0.3 drop in both box AP and mask AP.

\subsection{Additional Analysis}
\begin{figure}[h]
\begin{center}
\includegraphics[width=.9\linewidth]{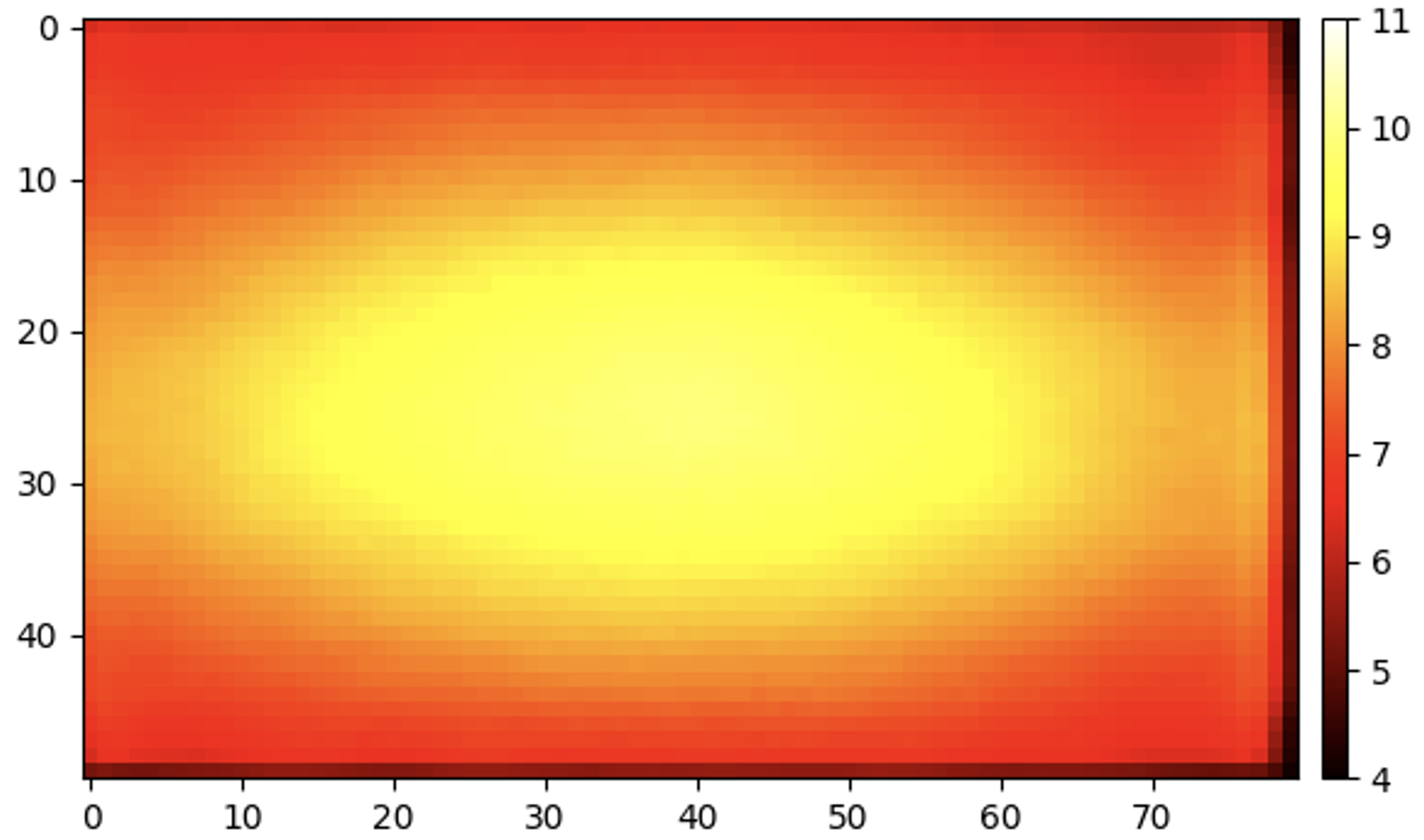}
\end{center}
\vspace{-5pt}

   \caption{Averaged token-usage heat map of SViT-T showing the number of active layers for each token position, averaged on COCO \cite{COCO} validation set.  The resolution is interpolated to 50 x 80 tokens for all images.}
   \vspace{-10pt}
\label{fig:heatmap_coco}
\end{figure}

\paragraph{Qualitative Results} We show qualitative results of the token pruning in SViT in Figure \ref{fig:qualitative_results}, and refer to supplementary material for more examples. The token-usage heatmap, created by quantifying the number of active layers for each token position, distinctly highlights not only the objects themselves but also their fine-grained contours. For example, the zebra's feet and the contour of the donuts stand out clearly against their respective backgrounds. In the case of background tokens, uniform areas such as the ground in the zebra image are more prone to be pruned, whereas textured backgrounds near objects are kept processed more. We also present the averaged token-usage heatmap on COCO validation set in Figure \ref{fig:heatmap_coco}. The heatmap reveals a higher frequency of token usage in the central regions of images, due to the common photographic tendency to place objects at the center.

\begin{figure}[h]
    \begin{center}
    \includegraphics[width=1.\linewidth]{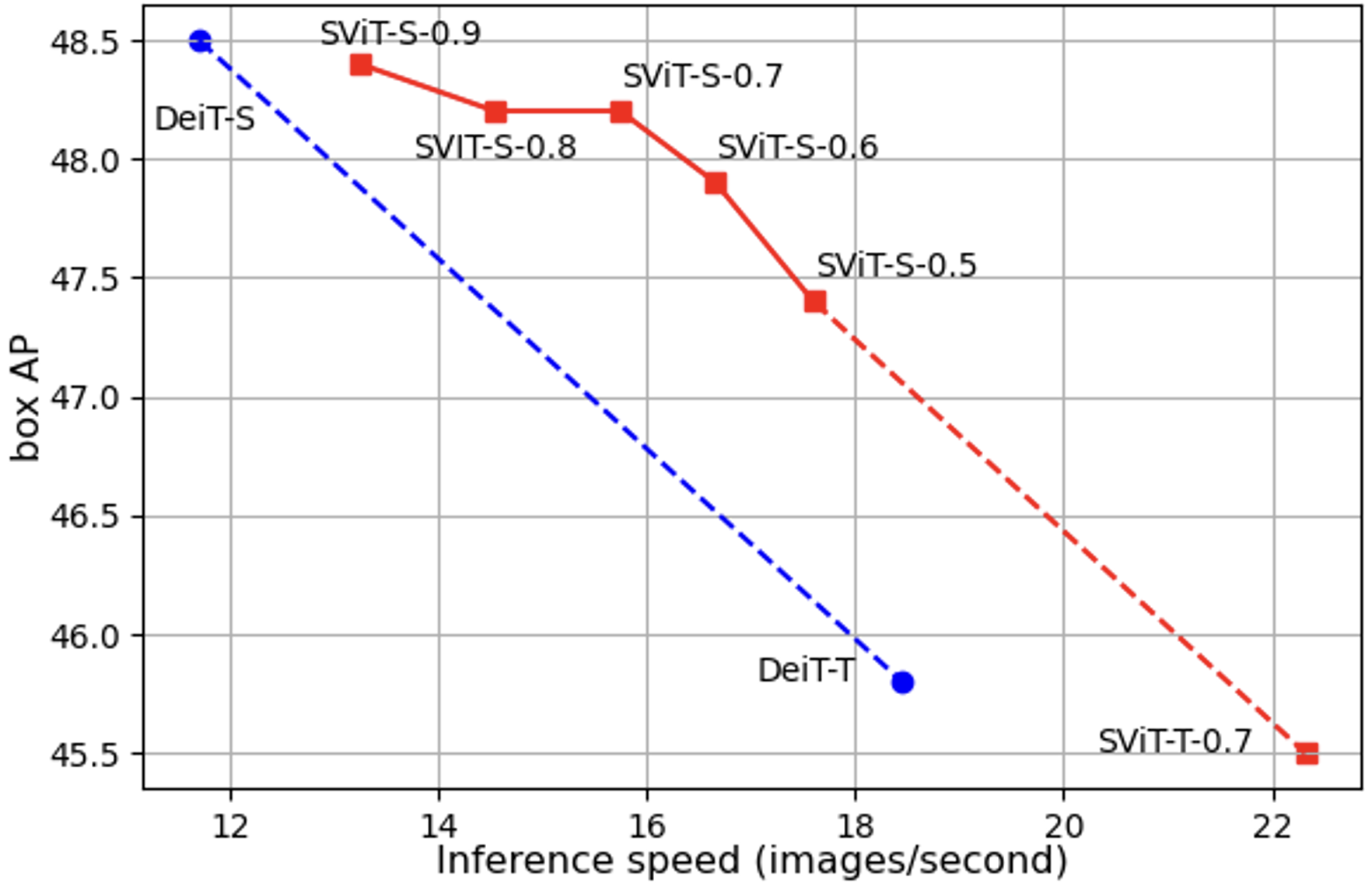}
    \end{center}
    \vspace{-5pt}
    
       \caption{Trade-off between speed and accuracy across various pruning rates in ViT-Adapter with dense DeiT and sparse SViT configurations.}
       \vspace{-10pt}
    \label{fig:model_scaling}
\end{figure}

\paragraph{Different pruning rates}
We adjust the pruning rate for SViT-S from the default 0.7 to \{0.5, 0.6, 0.8, 0.9\} and plot the mAP vs. pruning rate in Figure \ref{fig:model_scaling}. As shown in the plot, SViT consistently achieves better speed-accuracy trade-off than DeiT. However, we observe noticeable AP drop when base pruning rate is as low as 0.6 or 0.5, due to too aggressive pruning rates in the last three ViT blocks, i.e., 0.216 and 0.125, which is consistent with findings from classification \cite{DynamicViT}.

\paragraph{Reactivation distribution} To further understand the behavior of reactivation across transformer layers, we plot the reactivation ratio at each layer averaged on COCO validation set in Figure \ref{subfig:reactivation_ratio}. In the scope of this paper, the reactivation ratio at a layer (cyan colored) is defined as the ratio of current pruned tokens that are reused in at least one later layer in the backbone. As shown in the plot, most pruned tokens in early layers are reused in later layers. This indicates that it is harmful to fully drop tokens in early layers, and the model chooses to recover them in succeeding layers to alleviate the loss. In deeper layers, although the pruning rate is higher, the reactivation ratio is not apparently increased, as it is more tolerant to drop tokens. We also observe that over 50\% of reactivated tokens are immediately reused in the succeeding layer. This observation aligns with the notion that the utility of a token diminishes if it remains unused for an extended period, given that feature characteristics often vary between deep and shallow layers.

\vspace{-10pt}
\paragraph{Reactivation areas}
In the previous section we analysed the reactivation ratio for different model layers, and here we show reactivation regions in images as visualized in Figure \ref{subfig:reactivation_example}. The token pruning is shown at middle layers of SViT. As anticipated, background tokens are predominantly not reactivated (white colored), while pruned tokens in interested objects, such as person, soccer and computers, are selectively reactivated (cyan colored). When faced with a high pruning rate that necessitates the temporary removal of tokens associated with objects of interest, the model strategically reactivates these tokens at later layers. This approach allows for a more expansive set of active tokens compared to scenarios where token reactivation is not an option.
\section{Limitations and Societal Impacts}
\paragraph{Limitations} The aim of our work is to bridge the gap of token pruning between classification and dense tasks for isotropic vision transformers. We do not focus on pyramidal vision transformers, nor on exploring better pruning rates. These topics are covered by some concurrent works and will be further studied in future works.
\vspace{-10pt}

\paragraph{Societal Impact} The fintuning of sparse pruning is conducted after the model is fully trained and will introduce some additional energy consumption for training. However, this cost can be amortized once the model is deployed with improved inference efficiency. The proposed method predicts pruned tokens based on learned statistics from the training dataset, any bias inherent in the training data will be mirrored in the pruning process, and may result in the model disregarding biased content and exacerbating fairness issues.
\section{Conclusions}
\label{sec:conclusions}

In this work, we revisit the designs of token pruning for vision transformers in the context of object detection and instance segmentation. We provide four insights that can enhance token pruning on dense tasks: the pruned tokens should not be removed but preserved in feature maps; reactivating pruned tokens at demand can boost model performance; a dynamic pruning rate is helpful on dense tasks; and a 2-layer MLP can be as effective as more complex gating networks. By incorporating these insights together, we present a token pruning method that outperforms prior state-of-the-arts by a significant margin and accelerates backbone inference by $\sim$46\% with negligible loss in accuracy. We hope these insights and encouraging results can inspire further research on ViT acceleration for dense prediction tasks beyond image classification.

\section{Acknowledgements}
This work was supported by the National Centre of Competence in Research (NCCR) Robotics (grant agreement No. 51NF40-185543) through the Swiss National Science Foundation (SNSF), and the European Research Council (ERC) under grant agreement No. 864042 (AGILEFLIGHT).

\newpage
\section*{Supplementary Material}

\section*{A \hspace{0.2cm} Results on ImageNet-1K classification}
\label{appx:D}
\paragraph{Setup} We also train and evaluate SViT-S on ImageNet-1K \cite{ImageNet1k}. We follow the training settings in DeiT \cite{DeiT} and initialize our model from public pre-trained weights of DeiT-S. We use an AdamW optimizer to train the our model for 30 epochs and set the learning rate as $\frac{\text{batchsize}}{512} \times$ 1e-5. The model is trained on a single machine with 4 V100 GPUs with a batch size of 1024.

\vspace{-10pt}

\paragraph{Results} We compare the throughput of SViT-S and the dense counterpart DeiT-S in Table \ref{tab:Throughput_Classification}. SViT achieves 47\% higher throughput than the dense counter part while only sacrificing -0.4\% accuracy, effectively improving the accuracy-speed trade off. We also compare SViT-S with other token pruning models in Table \ref{tab:Performance_ImageNet1k}. Although SViT is not originally targeted at classification tasks, it outperforms all models that use gating modules (DynamicViT \cite{DynamicViT}, SPViT \cite{SPViT}, AdaViT \cite{AdaViT}), the models using special pruning techniques such as adaptive computation time \cite{ACT} (A-ViT \cite{A-ViT}) and reinforcement learning (IA-RED2 \cite{IA-RED2}), and a model that uses class token's attention (Evo-ViT \cite{EvoViT}).
However, when it comes to classification, EViT and ATS demonstrate superior performance over SViT. This is primarily due to their utilization of the class token, a feature specifically designed for the classification task.
\begin{table}[h]
    \small
    \begin{center}
    \begin{tabular}{l c c c}
    \hline
    Model & Top-1 Accuracy & GFLOPS & $\text{images}/s$\\
    \hline
    DeiT-S \cite{DeiT} & 79.8 & 4.6 & 1524 \\
    SViT-S & 79.4 & 3.0 & 2246 \\ 
    
    \hline
    \end{tabular}
    \end{center}
    \vspace{-5pt}
    \caption{Model Performance of DeiT-S and SViT-S. Throughput is measured on a single A100 GPU with batch size 512.}
    \vspace{-5pt}
    \label{tab:Throughput_Classification}
\end{table}

\begin{table}[h]
    \small
    \begin{center}
    \begin{tabular}{l l c c}
    \hline
    Model & epochs & GFLOPS & Top-1 Acc(\%)\\
    \hline\hline
    DeiT-S \cite{DeiT}& - & 4.6 & 79.8\\
    \hline
    DynamicViT $\ddagger$ \cite{DynamicViT} & 30 & 3.0 & 79.3 (-0.5)\\ 
    EViT \cite{evit}& 30 & 3.0 & 79.5 (-0.3)\\ 
    Evo-ViT \cite{EvoViT}& 300 * & 3.0 & 79.4 (-0.4)\\
    Evo-ViT  \cite{EvoViT}& 30 $\dagger$ & 3.0 & 79.2 (-0.6) \\
    A-ViT \cite{A-ViT}& 100 & 3.6 & 78.6 (-1.3)\\
    ATS \cite{ATS} & 30 & 3.0 & 79.7 (-0.1) \\
    AdaViT \cite{AdaViT}& 150 & 2.3 & 77.3 (-2.5) \\
    IA-RED2 \cite{IA-RED2}& 90 & 3.2 & 79.1 (-0.7) \\
    SPViT $\ddagger$  \cite{SPViT}& 60  & 2.7 & 79.3 (-0.5) \\
    SViT (Ours) & 30 & 3.0 & 79.4 (-0.4) \\
    \hline
    \end{tabular}
    \end{center}
    \vspace{-5pt}
    \caption{Model performance on ImageNet-1K. * means training from scratch. $\dagger$ indicates experiments trained by us. $\ddagger$ uses additional knowledge distillation. Note that EViT, Evo-ViT and ATS depend on the class token, designed specifically for classification, to help improve their performance. On the other hand, SViT targets more general tasks and does not rely on the attention map of the class token for token pruning.}
    \label{tab:Performance_ImageNet1k}
    \vspace{10pt}
\end{table}

\section*{B \hspace{0.2cm} Influence of Batch Size on Throughput}
It is not straightforward to do batch inference with different number of tokens per image, as the tensor cannot be easily arranged in a regular shape, therefore, we use \textit{pytorch} \textit{nested tensor}\footnote {\url{https://pytorch.org/docs/1.13/nested.html}} to efficiently process the varying-length sequences of tokens. The tokens to be processed are first gathered into a nested tensor, then passed to a ViT block constructed with \textit{nested tensor} operations, and finally un-nested and scattered back to the feature map.

\begin{figure}[h]
\begin{center}
\includegraphics[width=1.\linewidth]{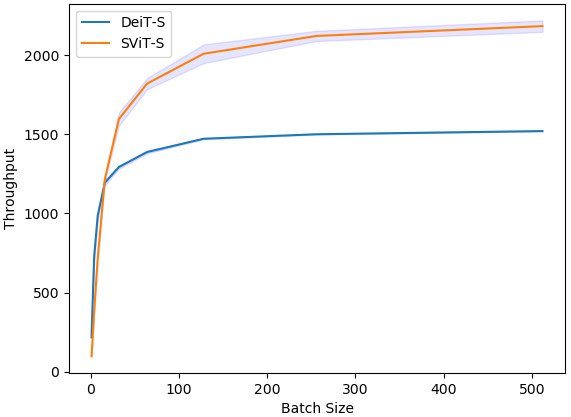}
\end{center}
\vspace{-5pt}
   \caption{Throughput vs. Batch Sizes of SViT-S and DeiT-S on ImageNet-1K.}
\label{fig:bs_throughput}
\end{figure}

We test the throughput of SViT and DeiT on ImageNet-1k for varying batch sizes as follows: for each batch size, we randomly fetch 30 batches from the validation set, and for each batch we run the inference for 50 times and take the average throughput as the speed for this batch. Then we calculate the mean and standard deviation over the speeds of the batches. As seen in Figure \ref{fig:bs_throughput}, proportional throughput gains can be obtained from increased batch sizes for SViT, which verifies that \textit{nested tensor} could be a promising way of handling the varying-sized tensors in the dynamic scenario. However, note that the \textit{nested tensor} is not fully developed and is still in a prototype stage, and leads to some overhead when the batch size is small. 

In the case of object detection and instance segmentation, we abstain from using \textit{nested tensors} due to the difficulties associated with creating a large batch size for high-resolution images. Consequently, we centered our efforts on inference with a batch size of 1 for these dense tasks. Future advancements in this technique, along with other related breakthroughs, may facilitate additional speedups. These improvements could be readily integrated into SViT, as previously demonstrated in classification tasks.
\label{appx:E}

\section*{C \hspace{0.2cm} ViT has different layer-wise attention}
\label{appx:B}
\begin{figure*}[h]
\begin{center}
\includegraphics[width=0.9\linewidth]{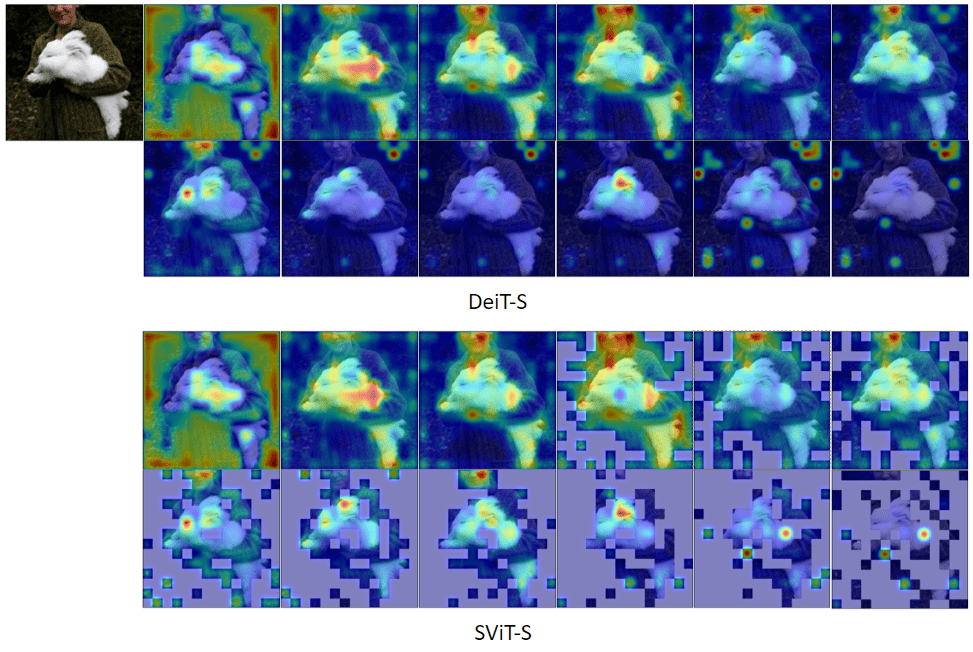}
\end{center}
\vspace{-10pt}
   \caption{The first row: the attention from the \nth{1} layer to the \nth{6} layer; the second row: the attention maps from the \nth{7} layer to the \nth{12} layer. Compared to the dense DeiT-S model, SViT-S keeps the most important features for each layer, especially the human face, the rabbit eyes, and the rabbit ear. Since ViT's attentions can be different across different layers, SViT does not keep the same tokens across different layers. The attention maps are visualized as the mean of the attention from class token's heads.  }
\label{fig:layer-wise attention}
\end{figure*}

Vision Transformers do not always attend to the same set of tokens, even for the important ones. An illustrated in the example in Figure \ref{fig:layer-wise attention}, the dense DeiT-S \cite{DeiT} first attends to the background tokens in the 1st layer, and then attends to joint regions of the human face and the rabbit in the next three layers. After that, the human face, rabbit eyes, and rabbit ears are attended in different layers, respectively. This inspired us to reactivate previously pruned tokens, as each layer can have its customized preference on tokens.

\section*{D \hspace{0.2cm} Discussion on Prior-Art Token Pruning Methods}
\label{appx:C}
Since the class token does not originally exist for ViT models on dense tasks, we append a randomly initialize a class token for attention-based token pruning models (EViT \cite{evit}, ATS \cite{ATS}, and Evo-ViT\cite{EvoViT}), which can help them prune tokens reasonably on dense tasks. 

\begin{figure*}[!h]
\begin{center}
\includegraphics[width=0.9\linewidth]{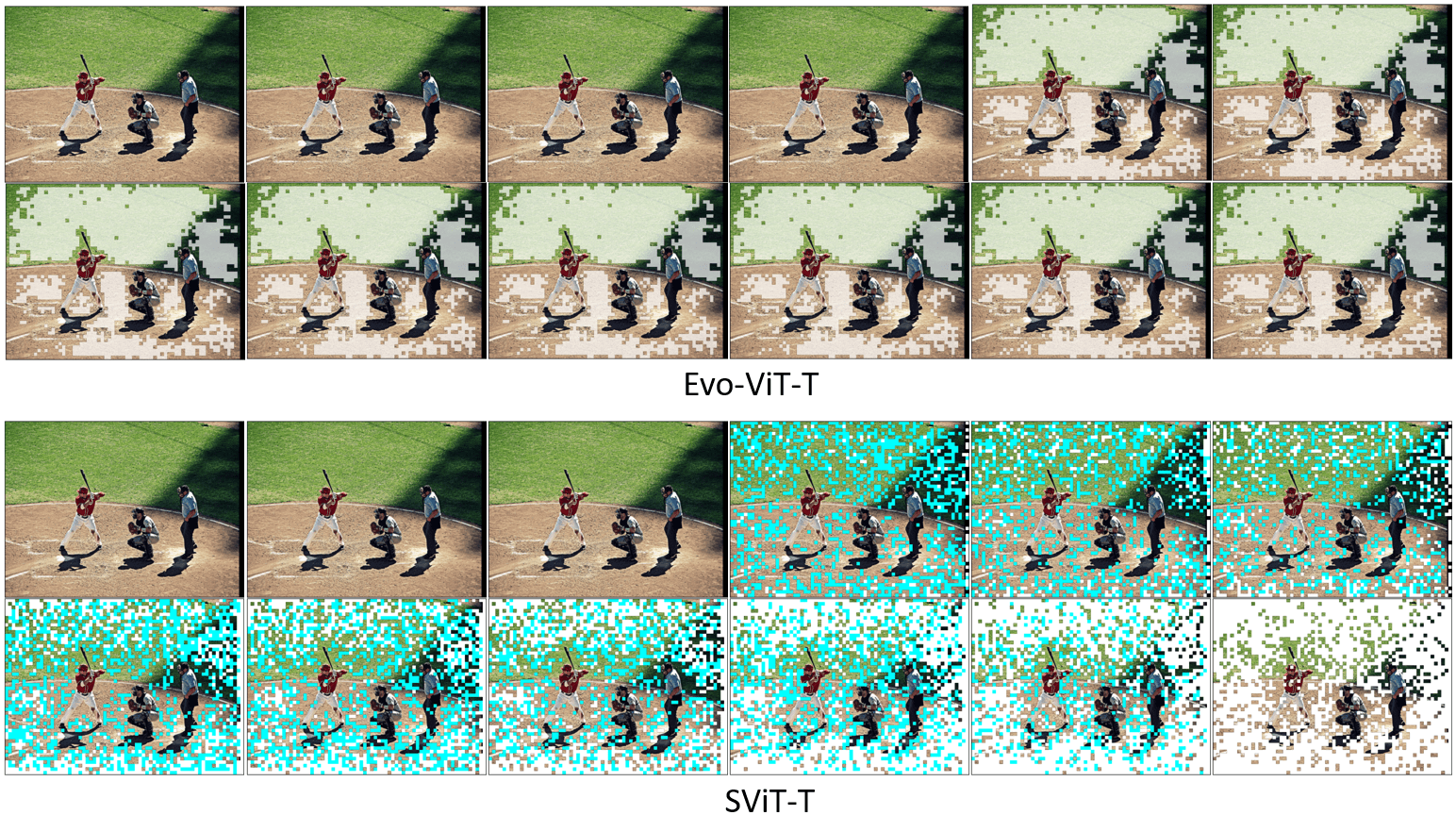}
\end{center}
\vspace{-5pt}
   \caption{Top: token pruning for Evo-ViT-T. Bottom: token pruning for SViT-T. Evo-ViT has the same keep ratio per layer, and tends to use the same set of tokens for all its pruning layers, so the selection largely depends on its first pruning layer, which may not be optimal. SViT can prune tokens independently for each layer, which is learned by the gating MLPs, and can reuse tokens according as needed.}
\label{fig:evo_not_reuse}
\end{figure*}

  Among these models, Evo-ViT is unique because it preserves pruned tokens in the feature map. However, it has a tendency to converge to a consistent set of tokens and thus does not reuse pruned tokens, as shown in Figure \ref{fig:evo_not_reuse}.
  We conjecture this is because Evo-ViT uses a moving average to update the attention scores of processed tokens, which in turn is used to select  tokens to be pruned. Since updating the scores is only done for processed tokens, the pruned tokens do not have a chance to change their scores, and thus causing the model to consistently use the same selection scheme.
As Figure \ref{fig:evo_not_reuse} illustrates, Evo-ViT consistently selects bottom-left tokens from its first layer onward, even though they are irrelevant background tokens. In contrast, our model can dynamically choose different tokens for each layer, and reuse important ones.

\balance
\section*{E \hspace{0.2cm} Qualitative Examples}
\label{appx:A}
\begin{figure*}[t!]
\begin{center}
\includegraphics[width=0.9\linewidth]{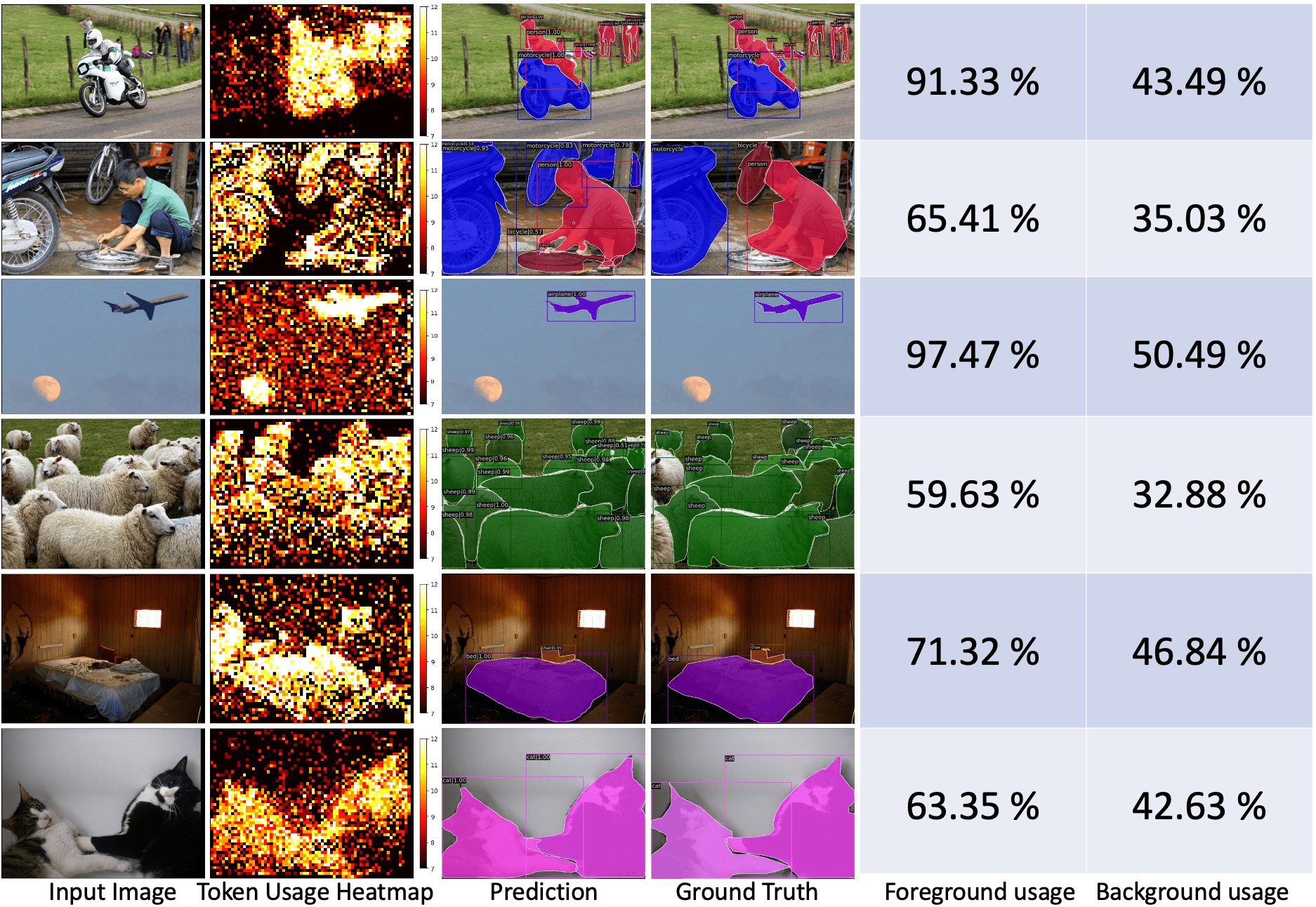}
\end{center}

   \caption{More visualizations of SViT-S on COCO validation set. The token usage heat map shows the number of used layers per token. The tokens are further split into two sets: foreground tokens and background tokens, and token usages are computed for them separately.}
\vspace{5in}

\label{fig:more_vis}
\end{figure*}

We provide more visualizations of SViT-S on COCO in Figure \ref{fig:more_vis}. To better understand the token selection of SViT, we split the tokens into two sets: object tokens and background tokens, according to the prediction mask.
Then we compute the token usage for them separately. The first observation is that SViT uses a larger ratio of foreground tokens than background tokens. When the proportion of the object in the image is small, the foreground token usage can be as high as 90\%. 
In cluttered images, such as the sheep example in the 4th row, not all foreground tokens are essential; less discriminative foreground tokens can still be pruned without affecting performance.

\clearpage

{\small
\bibliographystyle{ieee_fullname}
\bibliography{egbib}
}

\onecolumn
\twocolumn

\end{document}